\def\year{2018}\relax
\documentclass[letterpaper]{article} 
\usepackage{aaai18}  
\usepackage{times}  
\usepackage{helvet}  
\usepackage{courier}  
\usepackage{url}  
\usepackage{graphicx}  
\usepackage{subfigure}
\usepackage{todonotes}

\usepackage{amsmath, amssymb, amsthm}
\usepackage[shortlabels,inline]{enumitem}
\usepackage{booktabs}

\newtheorem{definition}{Definition}

\DeclareMathOperator*{\argmax}{arg\,\!max}

\nocopyright

\begin{document}
%
\title{Active Goal Recognition}
\author{Christopher Amato \and Andrea Baisero\\
Khoury College of Computer Sciences\\
Northeastern University\\
Boston, MA 02115
}
\maketitle

\begin{abstract}
To coordinate with other systems, agents must be able to determine what the
systems are currently doing and predict what they will be doing in the
future---plan and goal recognition.  There are many methods for plan and goal
recognition, but they assume a passive observer that continually monitors the
target system.  Real-world domains, where information gathering has a cost
(e.g., moving a camera or a robot, or time taken away from another task), will
often require a more active observer.  We propose to combine goal recognition
with other observer tasks in order to obtain \emph{active goal recognition}
(AGR).  We discuss this problem and provide a model and preliminary
experimental results for one form of this composite problem. As expected, the
results show that optimal behavior in AGR problems balance information
gathering with other actions (e.g., task completion) such as to achieve all
tasks jointly and efficiently.  We hope that our formulation opens the door for
extensive further research on this interesting and realistic problem. 
\end{abstract}

\section{Introduction}

AI methods are now being employed in a wide range of products and settings,
from thermostats to robots for domestic, manufacturing and military
applications.  Such autonomous systems usually need to coordinate with other
systems (e.g., sensors, robots, autonomous cars, people), which requires the
ability to determine what the other systems are currently doing and predict
what they will be doing in the future.  This task is called plan and goal
recognition.

Many methods already exist for plan and goal recognition (e.g.,
\cite{kautz91formal,rao94,charniak93,goldman99,Ramirez10,Ramirez11,Fern10,geib15acs};
see \cite{PlanRecognition14} for a recent  survey).  However, these methods
assume that a passive observer continually observes the target (possibly
missing some observation data), that is no cost to acquire observations, and
that the observer has no other tasks to complete.

These assumptions fall short in real-world scenarios (e.g., assistive robotics
at home or in public) where robots have their own tasks to carry out, and the
recognition of others' goals and plans must be incorporated into their overall
behavior.  For example, consider a team of robots assisting a disabled or
elderly person: the robots would be tasked with activities such as fetching
items and preparing meals, while also opening doors or otherwise escorting the
person. 
As a result, the robots will need to balance completion of their own tasks with
information gathering about the behavior of the other (target) agent. 

Current goal recognition methods do not address this \textit{active} goal
recognition problem.  While there has been some work on integrating plan
recognition into human-machine interfaces~\cite{kamar13,Freedman17} or
more generally addressing observation queries~\cite{mirsky16}, it has been
limited to utilizing query actions in service of recognition, rather than in an
attempt to reason about recognition through general tasks. 

In contrast, real-world domains will involve agents interacting with other
agents in complex ways.  For example, manufacturing and disaster response
will likely consist of different people and robots conducting overlapping tasks
in a shared space.  Similarly, agents and robots of different make and capacity
will be deployed in search and rescue operations in disaster areas, forming an
ad hoc team that requires collaboration without prior
coordination~\cite{huber-phd,aij13peter}.

Instead of considering a passive and fixed observer, 
we propose \emph{active} goal recognition for combining the observer's planning
problem with goal recognition to balance information gathering with task
completion for an observer agent and target agent. The active goal recognition
problem is very general, considering costs for observing the target that could
be based on costs of performing observation actions or missed opportunities
from not completing other tasks. The observer and target could be deterministic
or stochastic and operate in fully or partially observable domains (or any
combination thereof). In our formulation, we assume there is a planning problem
for the target as well as a planning problem for the observer that includes
knowledge and reasoning about the target's goal.

\begin{figure}[t]
\centering
\includegraphics[width=0.55\linewidth]{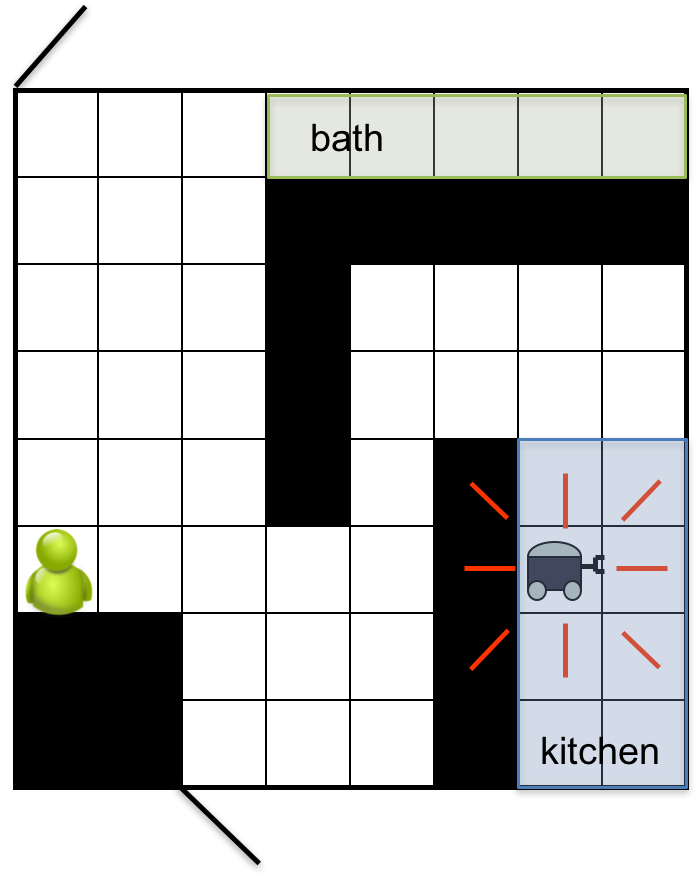}
\vspace{-7pt}
\caption{A depiction of  active goal recognition, where the robot has to balance information gathering using limited sensors with task completion (e.g., gathering items or cleaning an area) to assist the human and complete tasks.}
\label{fig:Domains}
\end{figure}

An example active goal recognition scenario is shown in
Figure~\ref{fig:Domains}.  Here, a single robot is tasked with cleaning the
bathroom and retrieving food from the kitchen, while also opening the proper
door for the person.
The robot must complete its tasks while predicting when and where the human
will leave.  This domain is a realistic example of assistive agent domains;  it
has the goal recognition problem of predicting the humans' goals as well as
a planning problem for deciding what to do and when.  This example is a simple
illustration of the class of agent interaction problems that must be solved,
and appropriate methods for active goal recognition will allow the autonomous
agents to properly balance information gathering with task completion in these
types of domains.

While the problem is very general, to make the discussion concrete, we describe
a version of the problem with deterministic action outcomes and both agents
acting in fully observable domains.  We also assume the target remains
partially observable to the observer unless the observer is in the proper
states or chooses the proper actions.  This research will be the first to
consider active goal recognition.  The goal is to extend the utility of goal
recognition, making it a viable task while also interacting with other agents

In the following section, we provide background on goal recognition and
planning.  We then describe our proposed active goal recognition problem and
representation.  We also present initial results from transforming our
deterministic and fully observable active goal recognition problems into
a partially-observable Markov decision process (POMDP) and using an
off-the-shelf solver to generate solutions.  Experiments are run on two domains
inspired by Figure~\ref{fig:Domains}. The results show that, as expected, it is
important to balance information gathering and other costs (e.g., task
completion).  We discuss the limitations of this simple representation and
solution methods as well as some proposed future work. 

\section{Background}

We first discuss goal recognition and then discuss planning using the POMDP
formulation. 

\subsection{Goal recognition}

Plan and goal recognition have been widely studied in the planning community
\cite{PlanRecognition14}.  Early methods assumed the existence of a plan
library and then attempted to determine the correct plan (and the correct goal)
from observations of the plan execution \cite{Kautz86,Lesh95,Huber94}.
Techniques for plan and goal recognition include Bayes nets  \cite{Charniak91},
hidden Markov models \cite{Bui03} and grammar parsing \cite{Geib09}.  In these
methods, goal recognition is treated as the process of matching observations to
elements of the given library.  This process, while effective, does not easily
admit planning or reasoning about which observations are needed, using familiar
planning methods.

More recently, Ram{\'\i}rez and Geffner have reformulated plan and goal
recognition as a planning problem, where it is assumed that the observed agent
(which we will call the \emph{target}) is acting to optimize its costs in
a known domain \cite{Ramirez09,Ramirez10,riabov2016plan}. In this case,
a classical planner is used to solve a planning problem for each of the
possible goals, while ensuring plans adhere to the observations and the known
initial state. These approaches are typically more flexible than methods that
use a plan library, as they dynamically generate plan hypotheses based on the
observations, rather than from a fixed library.

More formally, a planning problem is defined as a tuple $\langle S, I, A,
G \rangle$, where $S$ is a set of states, $I$ is the initial state, $A$ is
a set of actions, and  $G$ is a set of goals.  A plan is a sequence of actions
which changes the state from the initial one to (hopefully) one of the goal
states.  If actions have costs, then the optimal plan is that which reaches
a goal while while minimizing the sum of action costs.

While the use of a planner in the recognition process is promising, the
reliance on classical planning---characterized by deterministic actions and
fully-observable states---is a limiting factor which only adds to the
assumption of a passive observer.
To remove these limitations, some extensions have been proposed.  For example,
for the case where the state for the target is still fully observable, but
transitions are stochastic, goal recognition has been performed over Markov
decision processes (MDPs)~\cite{Baker09} and POMDPs~\cite{Ramirez11}. Here, the
goal recognition problem is reformulated as a planning problem, but instead of
solving multiple classical planning problems, multiple MDPs are solved.

While the goal recognition methods perform well in a range of domains, they
assume \emph{passive observation of the target}. In many real-world domains,
such as the robotic domains depicted in Figure \ref{fig:Domains}, the observers
will be mobile and have other tasks to complete besides watching the target.
\emph{Active goal recognition} requires a deeper integration of planning and
goal recognition.  This work will apply in the deterministic, fully observable
case as well as stochastic, partially observable and multi-agent cases, making
it a very general framework for interaction with other agents. 

\subsection{Partially Observable Planning}

Our active goal recognition problem can be formulated as a planning problem
with partial observability over the target agent.  We will discuss this
formulation in more detail in the next section, and first discuss general
planning under uncertainty and partial observability with POMDPs. 

A partially observable Markov decision process (POMDP) \cite{Kaelbling98}
represents a planning problem where an agent operates under uncertainty based
on partial views of the world, and with the plan execution unfolding over time.
At each time step, the agent receives some observation about the state of the
system and then chooses an action which yields an immediate reward and causes
the system to transition stochastically.  Because the state is not directly
observed, it is usually beneficial for the agent to remember the observation
history in order to improve its estimate over the current state. The belief
state (a probability distribution over the state) is a sufficient statistic of
the observation history that can be updated at each step based on the previous
belief state, the taken action and the consequent observation.  The agent
continues seeing observations and choosing actions until a given problem
horizon has elapsed (or forever in the infinite-horizon case). 

Formally, a  POMDP is defined by tuple $ \langle  S, A, T, R, Z, O, \mathcal{H}\rangle $,
where:
\begin{itemize}\setlength\itemsep{0mm}
\item 
$S$ is a finite set of states with designated initial state
distribution $b_0$;
\item  
$A$ is set of actions;
\item  
$T$ is a state transition probability function, $T\colon S \times A \times S \to [0,1]$, that specifies the
probability of transitioning from state $s \in S$ to $s' \in S$ when action $a \in A$ is taken (i.e., $T(s,a,s') = \Pr(s' \mid  a,s)$);
\item 
 $R$ is a reward function $R\colon S \times A \to \mathbb{R}$, the immediate reward for being in state $s \in S$ and taking 
action $a \in A$;
\item
 $Z$ is a  set of observations;
\item
  $O$ is an observation probability function $O\colon Z \times A \times S \to [0,1]$, the
probability of  seeing  observation $o \in Z$  given 
action $a \in A$  was taken which results in state $s' \in S$ (i.e.,  $O(o,a,s') = \Pr(o \mid a,s')$);
\item 
and $\mathcal{H}$ is the number of steps until termination, called the horizon.
\end{itemize}

A solution to a POMDP is a \emph{policy}, $\pi$. The policy can map observation histories to actions, $\pi\colon H \to A$, where $H$ is the set of observation histories, $h = \{o_i^1, \ldots, o_i^t \}$, up to time step, $t$ or, more concisely belief states to actions $\pi\colon B \to A$, where $B$ represents the set of distributions over states $S$. 
 
The value of a policy, $\pi$, from state $s$ is
$ V^\pi(s)=\mathop{\rm E} \left[ \sum_{t=0}^{h-1} \gamma^tR(\vec a^t, s^t) \mid s,\pi \right]$,
which represents the expected value of the immediate rewards summed for each step of the problem, given the action prescribed by the policy. 
In the infinite-horizon case ($\mathcal{H}=\infty$) the discount factor $\gamma \in [0,1)$ is included to obtain a finite sum. An  \emph{optimal policy} beginning at state $s$ is 
$\pi^*(s) = \argmax_{\pi} V^{\pi}(s)$. The goal is to find an optimal policy beginning at some initial belief state, $b_0$.

POMDPs have been extensively studied and many solution methods exist (e.g., for a small sample \cite{Kaelbling98,Shani12,Kurniawati08,Ross08,Silver10}). POMDPs are general models for representing problems with state uncertainty and  (possibly) stochastic outcomes. Generic solvers have made great strides at solving large problems in recent years \cite{Silver10,Ye17} and specialized solvers can be developed that take advantage of special structure in problem classes. 

\section{Active Goal Recognition}

A rich set of domains consist of a single agent conducting tasks that are difficult to complete (an example is shown in Figure \ref{fig:Domains}).
One author encounters these problems each day when he attempts to get three small children ready for school:  there is no way to get himself ready while making sure they are dressed, have breakfast, brush their teeth and deal with the inevitable catastrophes that arise. They somehow manage to get out of the door, but it is often not an enjoyable experience for any involved (and they are sometimes missing important things like shoes or lunches).  Having an autonomous agent to assist with some of these tasks would be extremely helpful. 
Of course, there are many other instances of similar problems such as having an autonomous wingman
in combat missions to protect and assist a human pilot \cite{Wingman}, a robot assisting a disabled person to navigate an environment, retrieve objects and complete tasks (e.g., opening doors or preparing meals as in Figure \ref{fig:Domains}) 
or a robot performing a search and rescue task with the help of another robot from a different manufacturer. 

An important piece of solving this problem is active goal recognition---recognizing the goal of the target agent while the observer agent is completing other tasks. For example, in a room with multiple doors, robots would need to determine when a disabled person is going to go out a door and which one. Some work has been done in person tracking and intent recognition (e.g., \cite{PlanRecognition14,Cohen08,Gockley07}), but improving the methods and integrating them with decision-making remains an open problem.

One version of the active goal recognition problem is defined more formally below, but like the passive goal recognition problem of Ram{\'\i}rez and Geffner, we propose transforming the active goal recognition problem into a series of planning problems. In particular, like the passive problem, we can assume the domain of the target is known (the states, actions, initial state and possible goals), and given a set of observed states for the target, planning problems can be solved to reach the possible goals. The difference in the active case is that the observer agent's observations of the target depend on the actions taken in its domain (which may be different than the target's domain). 

We now sketch an overview of our proposed approach and a preliminary model.
The active goal recognition problem considers the planning problem of the observer and incorporates observation actions as well as the target's goals into that problem. These observation actions can only be executed based on preconditions being true (such as being within visual range of the target). The goals of the observer are augmented to also include prediction of the target's true goal (e.g., moving to the target's goal location or a more generic prediction action). Therefore, the observer agent is trying to reach it's own goal as well as correctly predict the chosen goal of the target (but observations of the target will be needed to perform this prediction well).
This augmented planning problem of the observer, which includes knowledge of the target's domain and observation actions is \emph{the active goal recognition problem}. By solving the augmented planning problem for goals of the target, we can predict the possible future states of the target and act to both complete the observer's planning problem and gain information about the target.

More formally, a sketch of a definition and model is:
\begin{definition}[Active Goal Recognition (AGR)]
Given a planning problem $P_O$ for the observer agent, and 
a planning domain $D_T$ as well as possible goals for the target agent $G_T$, construct a new planning problem $P_{AGR}$.
\end{definition}
Note that a domain is a planning problem without a (known) goal and the domains for the target and observer do not have to be the same, but they have to be known (e.g., maybe the agent just needs to predict the correct goal without acting in the same world as the target). 
Given (classical) planning representations of $P_O$ and $D_T$, the planing problem,  $P_{AGR}$, can be constructed  as a tuple $\langle S, I, A, G \rangle$, where
\begin{itemize}
\item $S_P \times S_T$ is a set of states for the observer and the target, 
\item $I_P$ is the initial state in the observer's planning problem,
\item $I_T$ is the initial state in the target,
\item $A_P \cup A_O \cup A_D$ is a set of actions for acting in the observer's planning problem, $A_P$, observing the target, $A_O$, and deciding on the target's goal,   $A_D$
\item $G_P$ is a set of goals in the observer's planning problem, and 
\item $G_T$ is a set of possible goals for the target. 
\end{itemize}

A solution to this planning problem is one that starts at the initial states of the observer and target and chooses actions that reach an augmented goal $G_{PT}$ (with lowest cost or highest reward). The augmented goal, $G_{PT}$, is a combination of the observer's and target's goal: satisfying the conditions of the observer's goal as well as predicting the target's goal. Prediction is  accomplished with, $A_D$, a prediction action that chooses the goal of the target.  $A_D$ may have preconditions requiring the observer to be in the target's goal location (e.g., in navigation problems), thereby ensuring the observer predicts correctly (costs can also be used to penalize incorrect predictions).
It may also be the case that $A_O$ is null and indirect observations are received through $A_P$ (e.g., the agent doesn't choose to observe, but it happens passively in certain states).  The planning problems for the target and observer could be deterministic or stochastic, fully observable or partially observable. The general AGR formulation makes no assumptions about these choices, but to make the problem concrete and simple, this paper focuses on the deterministic, fully observable case. 

For example, consider the problem in Figure \ref{fig:Domains}. Here, the robot and target (human) have similar, but slightly different domains. The target's domain is assumed to be just a navigation problem to one of the two doors. Of course, the target is a human, so it may conduct other tasks along the way or take suboptimal paths. The robot's domain consists of not just navigation, but actions for cleaning and picking up and dropping items (and the corresponding states). The goal for the robot in its original planning problem, $P_O$, consist of the bathroom being clean and the food item being in its gripper. The observation actions are null in this case and it is assumed that when the robot is in line of sight of the human, it can observe the human's location. The augmented goal consists of the original goal from $P_O$ as well as prediction of the target's goal, which in this case requires navigation to the predicted door location. Costs could vary, but for simplicity, we could assume each action costs 1 until the goal is reached. 

Unfortunately, this problem is no longer fully observable (because the state of the target is not known fully). As such, this problem can be thought of as a contingent planning problem or more generally as a POMDP. Next, we present a POMDP model, which we use for our experiments. 
The model and solutions serve as a proof of concept for the active goal recognition problem and future work can explore other methods (such as using contingent planning methods). 

\subsection{POMDP Representation}

\newcommand\Sa{S_P}
\newcommand\sa{s_P}
\newcommand\ssa{s'_P}
\newcommand\Aa{A_P}
\newcommand\Ta{T_P}
\newcommand\Ra{R_P}
\newcommand\Za{Z_P}
\newcommand\Ha{\mathcal{H}_P}

\newcommand\St{S_T}
\newcommand\Sto{S_T^o}
\newcommand\Stno{S_T^{\bar{o}}}
\newcommand\st{s_T}
\newcommand\sst{s'_T}
\newcommand\sto{s_T^o}
\newcommand\ssto{s_T^{\prime o}}
\newcommand\stno{s_T^{\bar{o}}}
\newcommand\sstno{s_T^{\prime\bar{o}}}
\newcommand\At{A_T}
\newcommand\at{a_T}
\newcommand\Tt{T_T}
\newcommand\Rt{R_T}
\newcommand\Ht{\mathcal{H}_T}
\newcommand\Gt{G_T}
\newcommand\gt{g_T}
\newcommand\ggt{g_T'}

\newcommand\Ao{{A_O}}
\newcommand\ao{{a_O}}
\newcommand\Apgr{{A_D}}
\newcommand\apgr{{a_D}}

\newcommand\Sagr{S}
\newcommand\Sagro{\Sagr^o}
\newcommand\sagr{s}
\newcommand\ssagr{s'}
\newcommand\sagrt{s^\dagger}
\newcommand\Aagr{A}
\newcommand\aagr{a}
\newcommand\Tagr{T}
\newcommand\Ragr{R}
\newcommand\Zagr{Z}
\newcommand\Oagr{O}
\newcommand\oagr{o}
\newcommand\oagrn{\oagr^\dagger}
\newcommand\oa{\oagr_P}
\newcommand\ot{\oagr_T}
\newcommand\Hagr{\mathcal{H}}

\newcommand\RR{{\mathbb{R}}}
\newcommand\Id{{\mathbb{I}}}

Given the planning problem for the observer agent $P_O$ and a planning domain
$D_T$ and goals $G_T$ for the target agent, we construct the AGR POMDP as the
tuple $\langle \Sagr, \Aagr, \Tagr, \Ragr, \Zagr, \Oagr, \Hagr \rangle$, where:
\begin{description}
  \item[The state space] $\Sagr=\Sa\times\St\times\Gt$ is the Cartesian product
    of observer states, target states and target goals, and states factorize as
    $\sagr=\left(\sa,\st,\gt\right)$;
  \item[The action space] $\Aagr=\Aa \cup \Ao \cup \Apgr$ is the union of
    actions available in the observer's domain, together with observation and
    decision actions;
  \item[The transition model] $\Tagr\colon\Sagr\times\Aagr\times\Sagr\to\RR$
    factorizes into the marginal observer and target transition models,
    respectively $\Ta$ and $\Tt$.  Let $\at$ indicate the action that the
    target makes in its own planning domain, then
    \begin{align}
      \Tagr\left(\sagr,\aagr,\ssagr\right) &= \Id\left[\gt=\ggt\right]
      \Tt\left(\st, \at, \sst\right) \nonumber \\
      &\phantom{=}\cdot \begin{cases}
        \Ta\left(\sa, \aagr, \ssa\right) & \text{if } \aagr\in\Aa \\
        \Id\left[\sa=\ssa\right] & \text{otherwise}
      \end{cases}
    \end{align}
    where $\Id$ the indicator function which maps $\texttt{True}\mapsto 1$ and
    $\texttt{False}\mapsto 0$.  As a result, the target's goal doesn't change
    and the observer's state only changes when taking a domain action (as
    opposed to an observation or decision action). 
  \item[The reward function] $\Ragr$ can be represented directly from the costs
    from the observer's task, observation costs and decision costs/rewards.
    That is, depending on the type of action the observer takes, it receives
    the corresponding cost (that may depend on its state and the state of the
    target):
    \begin{align}
      \Ragr(\sagr, \aagr) =& {-}\Id\left[\aagr\in\Aa\right] c_P(\sa, \aagr)
      \nonumber \\
      & {-}\Id\left[\aagr\in\Ao\right] c_O(\sa,\st,\aagr) \nonumber \\
      & {-}\Id\left[\aagr\in\Apgr\right] c_D(\sa,\st,\aagr) \,.
    \end{align}
  \item[The observation space] $\Zagr$ factorizes into an fully observable
    component for the observer's own state and a noiseless, but partially
    observable target state component.  Assume that the target state space
    factorizes in an observable component and a complementary unobservable
    component $\St = \Sto \times \Stno$, and let $\oagrn$ indicate that no
    target-state observation is made, then $\Zagr = \Sa \times \left(\Sto \cup
    \left\{\oagrn\right\}\right)$, and an observation $\oagr\in\Zagr$
    decomposes into $\oagr = \left(\oa, \ot\right)$. As such, the observer will be able to fully observe some target states and never observe others (as described next);

  \item[The observation model] $\Oagr$ is deterministic and target-state
    observations may be obtained explicitly when an observation action
    $\aagr\in\Ao$ is made, or implicitly when the joint observer-target state
    is in a subset $\left(\ssa,\sst\right)\in\Sagro\subseteq\Sa\times\St$ which
    warrants a target observation (e.g., when the agents are within
    line-of-sight). Therefore, the observer will perfectly observe its own
    state and may get a perfect observation of the target state if the
    observation action allows it or the observer is in a corresponding state:
    \begin{align}
      \Oagr(\oagr, \aagr, \ssagr) &= \Id\left[\oa=\ssa\right] \nonumber \\
      &\phantom{=}\cdot \begin{cases}
        \Id\left[\ot=\ssto\right] & \text{if\ } \aagr\in\Ao \text{\ or\
        } \left(\ssa,\sst\right)\in\Sagro \,, \\
        \Id\left[\ot=\oagrn\right] & \text{otherwise} \,;
      \end{cases}
    \end{align}
  \item[The horizon] $\mathcal{H}$ is set based on the planning problem. 
\end{description}
Of course, POMDPs can represent more general forms of AGR problems that include
stochasticity and consider more general observation, transition and reward
models.  Nevertheless, we believe this AGR POMDP model balances expressivity
and problem structure. 

\section{Experiments}

\newcommand\reward{{\mathcal{R}}}
\newcommand\return{{\mathcal{G}}}
\newcommand\belief{{\mathcal{B}}}
\newcommand\entropy{{\mathbb{H}}}

\newcommand\policyAGR{{\pi^*}}
\newcommand\policyLBT{{\pi_T}}
\newcommand\policyLBA{{\pi_A}}
\newcommand\policyUB{{\pi_\triangle}}

\newcommand\pos{{P}}
\newcommand\post{\pos_\text{task}}
\newcommand\posw{\pos_\text{work}}
\newcommand\powerset{\mathcal{P}}

\newcommand\Aidle{{\texttt{A\_idle}}}
\newcommand\Awork{{\texttt{A\_work}}}
\newcommand\Aobs{{\texttt{A\_obs}}}
\newcommand\Aopen{{\texttt{A\_open($p$)}}}
\newcommand\Aleft{{\texttt{A\_left}}}
\newcommand\Aright{{\texttt{A\_right}}}
\newcommand\Aup{{\texttt{A\_up}}}
\newcommand\Adown{{\texttt{A\_down}}}
\newcommand\Amove{{\texttt{A\_\{left,right,up,down\}}}}
\newcommand\Ahelp{{\texttt{A\_help}}}

\newcommand\Onull{{\texttt{O\_null}}}
\newcommand\Opos{{\texttt{O\_pos($p$)}}}

\begin{figure}[t]
  \centering
  \subfigure[Corridor Domain]{
  \includegraphics[width=.8\linewidth]{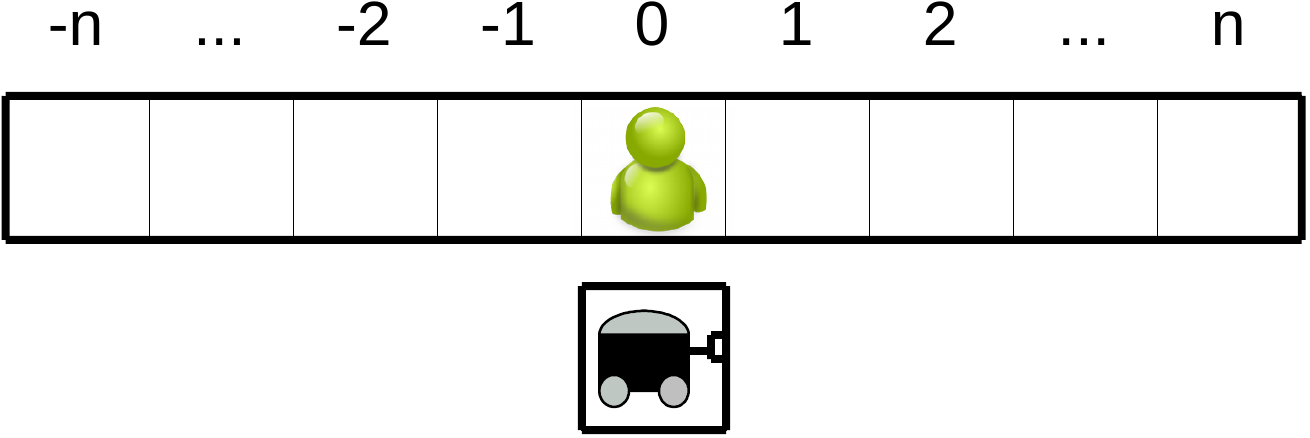} \label{fig:lindomain} }
  \hspace*{-.5cm}
  \subfigure[Map Domain]{ \includegraphics[width=.5\linewidth]{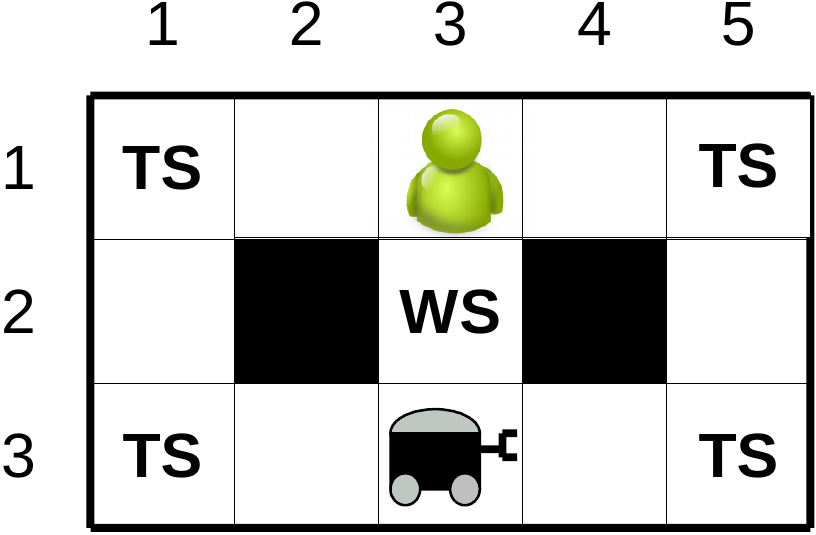}
  \label{fig:mapdomain} }
  \caption{AGR domains used in our evaluation.  Above, a corridor domain with
  size parameter $n=10$.  Below, a map domain with four task stations and an
occluded work station.}
  \label{fig:domains}
\end{figure}

We implement two prototypical AGR domains (Figure~\ref{fig:domains}): An
abstract \emph{corridor} domain and a more concrete \emph{map} domain. In both
domains, the observer must predict the goal of the target, while also
performing other work.  In the corridor domain, the observer and target do not
share the same working environment, and target-state observations can only be
obtained by making explicit observation actions.  In contrast, observer and
target in the map domain move in the same shared environment, and target-state
observations can only be implicitly obtained as a function of the joint state
rather than via an ad hoc action.

\subsubsection{Corridor Domain}

In this domain, depicted in Figure~\ref{fig:lindomain}, the target operates in
corridor with $10$ positions to the right and $10$ positions to the left ($21$
total positions).  Each space contains a door which can only be opened remotely
by the observer, and the target's goal is to walk through a randomly selected
goal door (known to the target only).  At each time-step, the target will move
towards the goal door, wait there if it is closed, or move to a terminal state
if it has been opened.  The observer is operating in a stateless multi-armed
bandit-like domain, and its own non-cooperative task is to either work or
remain idle.

Let $\pos$ be the set of all positions $\pos=\left\{-10,\dots,10\right\}$.  The
observer-, target-, and goal- states consist respectively of a singleton
$\Sa=\{0\}$, the set of positions augmented with a terminal state
$\St=\pos\cup\{\sagrt\}$, and a the set of all positions $\Gt=\pos$.  The
action space consists of actions for the observer planning problem
$\Aa=\{\Aidle,\Awork\}$, an explicit observation action
$\Ao=\left\{\Aobs\right\}$, and goal decision actions
$\Apgr=\{\Aopen\}_{p\in\pos}$.  The observable component of the target-state is
the target's position $\Sto=\pos$, while no direct observations can be made
about its goal.  There are no implicit observations, hence $\Sagro=\{\}$.  The
rewards are as follows:
\begin{description}[labelindent=\parindent,font=\normalfont]
  \item[\Aidle:]  Does nothing, obtains $\reward=0$;
  \item[\Awork:]  Performs work, obtains $\reward=10$;
  \item[\Aobs:]  Observes target position, obtains $\reward=-2$;
  \item[\Aopen:]  Represents the goal recognition action by attempting to open
    a door for the target, obtains $\reward=100$ if the correct door is chosen
    and the target is at the correct cell, otherwise $\reward=-100$.
\end{description}

Overall, the state-, action-, and observation- space sizes for this domain are
$|\Sagr|=462$, $|\Aagr|=24$, and $|\Zagr|=23$.

\subsubsection{Map Domain}

This domain (Figure~\ref{fig:mapdomain}) consists of a room containing walls,
task stations (representing the target task) and working stations (representing
the observer task).  The target task is to visit all task stations in any order
except for the last one---the goal task station---which encodes the target goal
and is randomly selected (known to the target only).  At each time step, the
target will move towards its closest next viable task station, until it reaches
the goal station where it will wait for the observer's help.  The observer's
non-cooperative task is to go perform work at a work station.

Let $\pos$ be the set of all positions which do not contain walls,
$\post\subseteq\pos$ be those with a task station, and $\posw\subseteq\pos$ be
those with a work station.  The observer-states consist of all observer
positions $\Sa=\pos$;  the target-states are the factorize as
$\St=\Sto\times\Stno$, where $\Sto=\pos\cup\{\sagrt\}$ is the observable set of
target positions enhanced with a terminal target state (which is reached after
the observer helps correctly) and $\Stno=\powerset\left(\post\right)$ is the
non-observable set of remaining task stations;  while the target goal is one of
the task stations $\Gt=\post$.  The action space consists of the actions
available for the observer planning problem
$\Aa=\{\Aidle,\Awork,\Aleft,\Aright,\Aup,\Adown\}$ and a goal decision action
$\Apgr=\{\Ahelp\}$ (there are no explicit observation actions, i.e.,
$\Ao=\{\}$).  The reward function is as follows:
\begin{description}[labelindent=\parindent,font=\normalfont]
  \item[\Aidle:]  Do nothing.  Obtains $\reward=0$;
  \item[\Awork:]  Attempts to work.  Obtains $\reward=5$ if the observer is at
    a working station, otherwise $\reward=-10$;
  \item[\Amove:]  Moves in the specified direction unless there is a wall.
    Obtains $\reward=-1$;
  \item[\Ahelp:]  Represents the goal recognition acion by attempting to help
    the target with the last task station.  Obtains $\reward=100$ if the target
    and observer are both at the goal task station and all task stations have
    been cleared, otherwise $\reward=-100$.
\end{description}

Overall, the state-, action- and observation- space sizes for this domain are
$|\Sagr|=6084$, $|\Aagr|=7$, and $|\Zagr|=195$.

\subsubsection{Lower and Upper Bounds}

Given an AGR domain, it is always viable (if suboptimal) to consider the
policies $\policyLBT$ and $\policyLBA$ which focus exclusively on the goal
recognition and observer task respectively (and ignore the other).  This
indicates that the value of the optimal policy $\policyAGR$ in the AGR domain
is bounded below by $V^\policyLBA \le V^\policyAGR$ and $V^\policyLBT \le
V^\policyAGR$, and inspires us to design two \emph{lower} bound (LB) variants
to the AGR domain.  The LB-A variant is constructed by applying a very strong
penalty ($\reward\gets\reward-10^6$) to the goal decision actions $\Apgr$, thus
inhibiting the observer's willingness to spend resources on gathering
information about the target's task.  Similarly, the LB-T variant is
constructed by applying the same penalty to a subset of the observer's own
planning problem actions $\Aa$ (in our domains, $\{\Awork\}$), thus ensuring
that the observer will not focus on its own task.

Furthermore, completing the goal recognition task is typically contingent on
the access to meaningful observations which are usually associated with an
acquisition cost.  This suggests that, if it were possible to have these
observations available at any time for no cost at all, the resulting optimal
policy $\policyUB$ would perform at least as well as the optimal AGR policy
$\policyAGR$ (i.e., $V^\policyAGR \le V^\policyUB$).  This inspires us to design
an \emph{upper} bound (UB) variant to the AGR domain.  The UB variant is
constructed by always giving target observations $\Sto$ to the observer,
regardless of current action or joint state.

\subsection{Results}

We compute policies for the corridor and map domains (and the respective LB and
UB variants) using the SARSOP solver \cite{Kurniawati08}.  For each domain and
variant thereof, we simulate 1000 episodes (to account for random goal assignments) and compute the set of returns for
each episode, and the current state-beliefs for each step.  SARSOP was chosen
because it is one of the most scalable optimal POMDP solvers, but other solvers
could also be used. 

\subsubsection{Controlled Goal-Belief Entropy Reduction}

Let $b_g$ be the goal-belief (i.e., the marginal state-belief obtained
summing over all non-goal components), and $\entropy_g$ be the normalized
goal-belief entropy:
\begin{align}
  b_g\left(g\right) &= \sum_{\sagr\in\Sagr\colon\gt = g} b\left(\sagr\right)
  \,, \\
  \entropy_g &= \frac{1}{\log|\Gt|} \sum_{g\in\Gt} b_g\left(g\right) \log
  \frac{1}{b_g\left(g\right)} \,.
\end{align}

\begin{figure*}[t]
  \centering
  \subfigure[Corridor Domain]{
  \includegraphics[width=.487\linewidth]{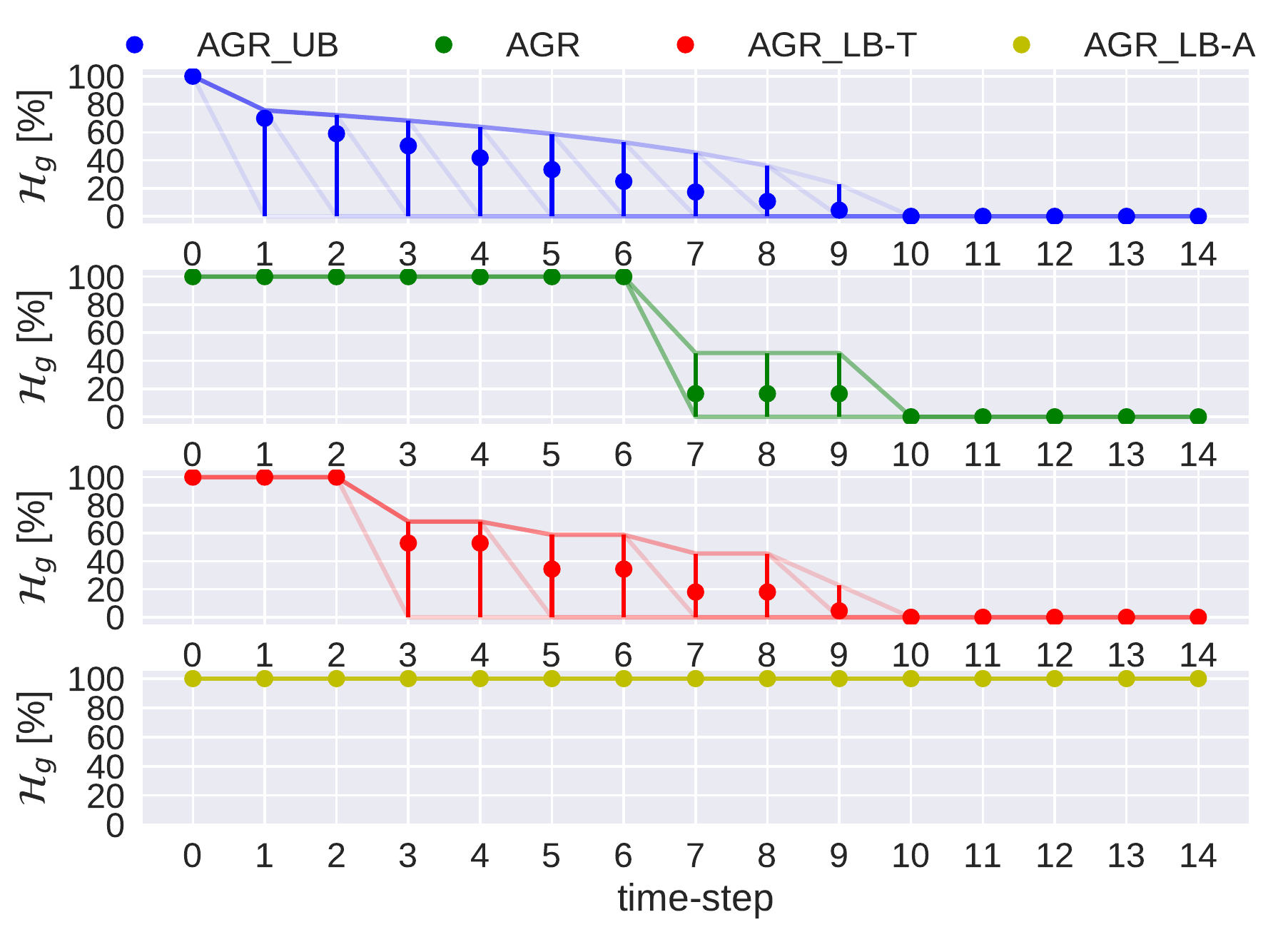} }
  \subfigure[Map Domain]{
  \includegraphics[width=.487\linewidth]{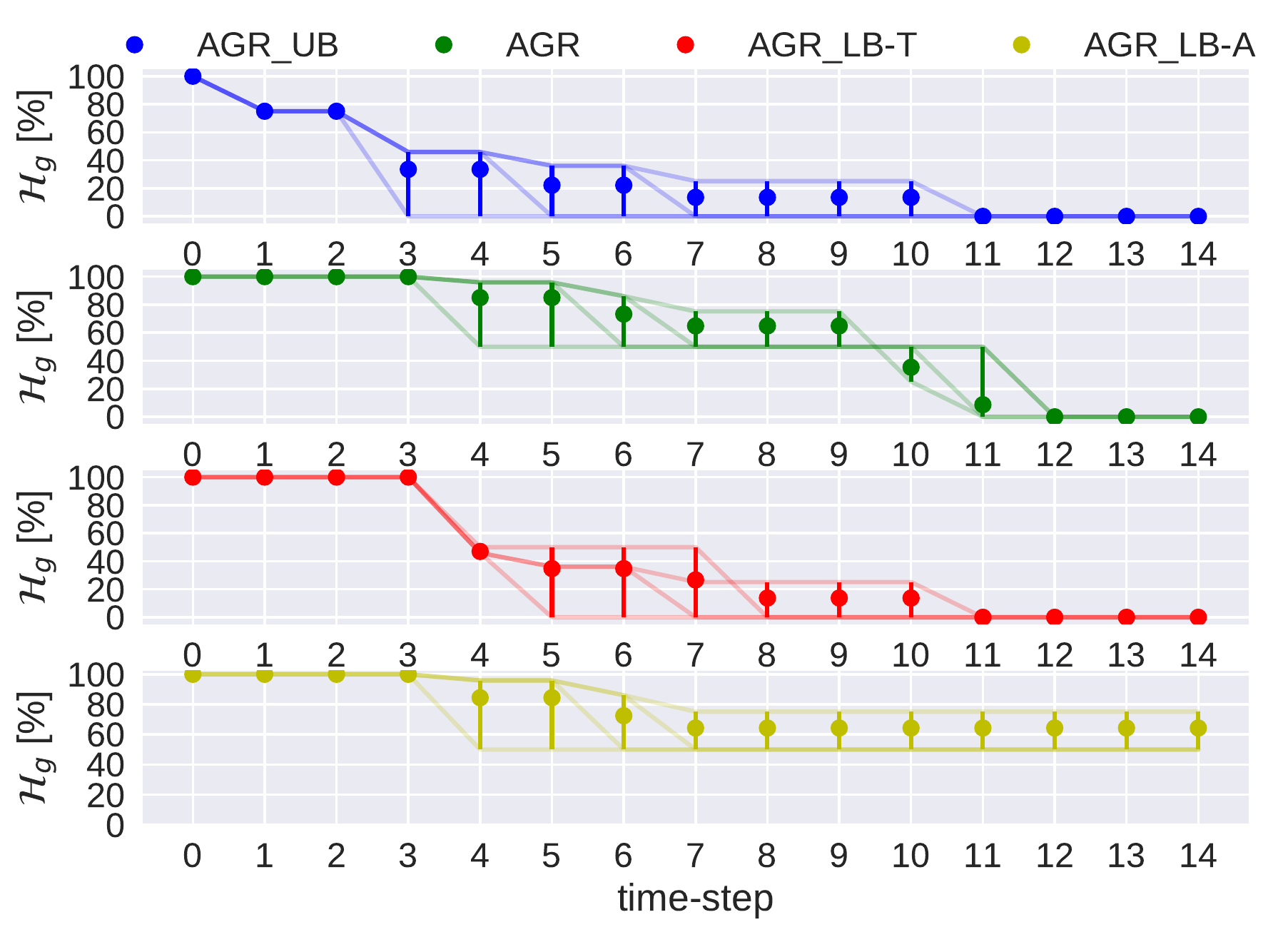} }
  \caption{Normalized goal-belief entropy $\entropy_g$ as a function of time.
    Dots represent mean entropies for a given time step; vertical lines
    represent the min-max range of entropies for that time step; and the
    horizontal or slanted lines indicate the change of entropy values across
  adjacent time-steps (the more opaque the line, the more frequent the
transition).}
  \label{fig:gbh}
\end{figure*}

Figure \ref{fig:gbh} shows the evolution of $\entropy_g$ during the course of
the sampled episodes, and a few observations can be made about it in each of
the 4 variants:
\begin{enumerate*}[a)]
  \item Given the static nature of the target goal in our domains, the mean
    $\entropy_g$ is non-increasing in time;
  \item because all observations are free in the UB variant, the respective
    goal entropy also shows the quickest average reduction;
  \item because the observer is prompted to focus only on the target task in
    the LB-T variant, the respective goal entropy also decreases relatively
    fast---albeit not necessarily as fast as in the UB case;
  \item because the observer is incentivized to parallelize the execution of
    both tasks in the AGR domain, the goal entropy decreases in average slower
    than in the UB and LB-T variants; and
  \item because the observer is uninterested in the target task, the LB-A
    variant is the only one where $\entropy_g$ is not guaranteed to converge
    towards zero.
\end{enumerate*}

Intuitively, the results show that the observer in the AGR domain tends to
delay information acquisition until it becomes useful (in expectation) to act
upon that information, yet not too much as to perform substantially
suboptimally w.r.t.\ the goal recognition task (i.e., the entropy decrease is
noticeably slower than in the UB and LB-T variants, but it still catches up
when it becomes possible to obtain complete certainty and guess the task
correctly).  As shown next, time is employed optimally by exploiting the
observation delays to perform its own task.

\subsubsection{Sample Returns}

Table \ref{tab:gkde} shows empirical statistics on the sample returns obtained
for each domain and bound variant.  As expected, the AGR method
performs better than both LBs, and worse than the UB.  In the previous section
we have shown that the certainty over the target goal in an AGR grows slowly
compared to the UB.  Here we see that this does not translate into much of
a loss in terms of expected returns, since the average AGR and UB performances
are within 1--2 standard deviations from each other.  The LB-A performance is
deterministic because the optimal strategy does not depend on the target state or goal.

We propose that such empirical statistics contain meaningful information about
the AGR design.  For instance, if the performance in an AGR domain and
respective UB is comparable, this should indicate that the observer's goal
recognition and own domain tasks can be efficiently parallelized---either
because observations are cheap, or because few observations at key moments are
sufficient, or because there are actions which move both tasks forward at the
same time.  On the other hand, if the AGR performance more strongly resembles
that of a LB, it may indicate that the rewards associated with the observer's
goal recognition and own domain tasks are too unbalanced, that one of the two
tasks is disregarded, and that a design revision of the AGR problem may be
necessary.

\begin{table}[t]
  \centering
  \begin{tabular}{lrrlrr}
    & \multicolumn{2}{c}{Corridor Domain} && \multicolumn{2}{c}{Map Domain} \\
    \toprule
    & mean & st.d. && mean & st.d. \\
    \cline{2-3} \cline{5-6} \rule{0pt}{3ex}
    UB & $223.1$ & $9.8$ && $104.6$ & $2.6$ \\
    AGR & $205.3$ & $7.9$ && $102.8$ & $2.4$ \\
    LB-A & $157.1$ & $0.0$ && $72.5$ & $0.0$ \\
    LB-T & $68.1$ & $11.9$ && $53.3$ & $1.6$ \\
    \bottomrule
  \end{tabular}
  \caption{Empirical mean and standard deviations for the sample returns
  obtained running the policy obtained by SARSOP in each AGR domain and LB/UB
variant.}
  \label{tab:gkde}
\end{table}

\subsubsection{Optimal Policies}

In the corridor domain, the observer spends the first few time steps performing
work rather than observing the target, due to the the fact that it is
relatively unlikely than observations at the beginning of an episode are
particularly informative about the target's goal.  A few time steps into the
episode, it performs the first observation, and then either opens the door, if
the target is found to be waiting, or goes back to work to check again after
another few time steps.  Comparatively, the LB-T behavior performs a similar
routine, except that it remains idle during the dead times rather than being
able to obtain more rewards by working.  In the ideal UB case, due to the
target positions always being observed, the observer is able to open the
correct door as soon as the target stops at a certain position, and uses the
spare time optimally by performing work.

In the map domain, the observer proceeds directly towards the nearby working
station and performs work until enough time has passed for the agent to be
approaching its goal task station.  At that point, the observer searches for
the target, helps, and then moves back to the work station.  The UB variant
performs a similar routine, except that it can (naturally) exploit the
additional information about the target's path to time its movement more
optimally:  While the \Ahelp\ actions in the AGR case are performed at time
steps 13--16 (depending on the goal task station), they are performed at time
steps 12--14 in the UB case.  This difference is small enough that the observer
decides to use its available time working rather than moving around trying to
get more precise information about the target's path and task.  As in the
corridor domain, the behavior in the LB-T case is similar to that of the AGR
domain, with the exception that the observer stays idle while waiting, rather
than profiting by working.

In both domains, the optimal LB-A behavior is straightforward---to work as
much as possible, while completing disregarding observations (explicit or
implicit) about the target task.

\section{Related Work}

Beyond the work already discussed, 
there has  been recent research on incorporating agent interaction into goal recognition problems.  For insstance, 
goal recognition has been formulated as a POMDP where the agent has uncertainty about the target's goals and chooses actions to assist them \cite{Fern10}. Also, the case has been considered 
 where the observer can assist the target to complete its task by solving a planning problem \cite{Freedman17}. 
In both cases, the observers sole goal is to watch the target and assist them. In contrast, we consider the more general case of \emph{active} observers that are also completing other tasks, balancing information gathering and task completion. 
We also consider multi-agent versions of the problem, unlike previous work.

There has also been a large amount of work on multi-agent planning (e.g., \cite{Durfee13,deWeerdt09,Vlassis07}).
Such methods have, at times, modeled teams of people and other agents, but they assume the people and other agents are also controllable and do not incorporate goal recognition.  
Ad hoc teamwork is similar, but assumes a set of agents comes together to jointly complete a task rather than an agent or team of agents assisting one or more agents or otherwise completing tasks in their presence ~\cite{huber-phd,aij13peter}.

Overall, as described above, research exists for goal recognition, but the work assumes passive and not active observers. 
This is an important gap in the literature. The problem in this paper represents a novel and the proposed methods allow an agent to reason about and coordinate with another agent in complex domains.

\section{Discussion}

In this paper, we used a general POMDP representation for the active goal recognition problem as well as an off-the-shelf solver. This was sufficient for the problems discussed, but both the model and solution methods could be substantially improved. POMDPs are general problem representations, but the problem representation can serve as inspiration to begin adding the appropriate assumptions and structure to the problem and developing solution methods that exploit this structure---finding the proper subclass and corresponding solution methods. 

In terms of the model, our AGR problem has a great deal of structure. The simple version assumes deterministic outcomes and agents that operate in fully observable domains. Furthermore, the goal of the target is assumed to be unknown, but does not change. As such, the belief can be factored into a number of components, with the observer's own state fully observable, but the target's is information partially observable. Furthermore, 
and as seen by the entropy analysis above, information is never lost as the belief over the target's goal continues to improve as more observations are seen. Stochastic and partially observable versions of AGR will also have this factored belief structure. This structure will allow specialized solution methods that are much more efficient than off-the-shelf methods such as the one used above. 

In terms of solution methods, many options are possible. Offline POMDP solvers (such as SARSOP \cite{Kurniawati08}) could be extended to take advantage of the special structure, but more promising may be extending online POMDP solvers  \cite{Ross08,Silver10}. Online solvers interleave planning and execution by planning for a single action only, executing that action, observing the outcome and then planning again. Online solvers are typically more scalable and more robust to changes. These online methods typically work by using a forward tree search (e.g., Monte Carlo tree search, MCTS \cite{Silver10}), which limits search to reachable beliefs rather than searching the entire state space. Therefore, online methods should be able to scale to very large state spaces, while also taking advantage of the factored states, actions and observations. 
In our experiments SARSAP could no longer solve versions of our domain with more than 6000 states. We expect online methods (such as those based on MCTS) to scale to problems that are orders of magnitude larger.

Other methods would also be able to take advantage of the special structure in
AGR. For instance, many methods exist for contingent planning in partially
observable domains (e.g., \cite{Maliah14,Muise14,Bonet17}). These methods may
be able to be directly applied to some versions of the AGR problem and variants
could be extended to choose (for instance) the most likely target goal and
replan when unexpected observations are seen (e.g., like \cite{Bonet11}).
Because of the information gathering structure of the belief, simpler (e.g.,
greedy) information-theoretic and decision-theoretic methods could also be used
that directly use the probability distribution over goals to explicitly reason
about information gathering and task completion.  The information-theoretic
case could consider planning actions that reduce the entropy of the
distribution over goals, while decision-theoretic methods could be greedy,
one-step methods that consider both the change in entropy and the action costs.
Future work on modeling and solving AGR problems as well as comparing these
solutions will show the strengths and weaknesses of the approaches and provide
a range of methods that fit with different types of domains.

\section{Conclusions}

Goal recognition is typically described as a passive task---an agent's only focus.  
We generalize this setting and provide an active goal
recognition (AGR) formulation in which the agent has other supplementary tasks
to execute alongside the original goal recognition task.  This AGR problem
represents a realistic model of interaction between an agent and another
system---ranging from a person to an autonomous car from a different
manufacturer---where both interactive and individual tasks come together.  Our
POMDP representation is one way of modeling such type of problems, but others
are possible.  We provide a tractable instance as well as preliminary results
showing the usefulness of this problem statement.  Empirical results show that,
rather than observing the target continuously, the optimal strategy in a well
crafted AGR domain involves the selection of key moments when observations
should be made such as to obtain important information about the target goal as
soon as if becomes relevant (rather than as soon as possible), and to have more
available time to continue performing other tasks.  These results represent
a proof of concept that AGR methods must balance information gathering with the
completion of other tasks, accounting for the respective costs and benefits.
We expect that further research efforts will be able to expand upon our
initial results, and introduce new exciting models and methods for
representing and solving this newly introduced class of problems.

{\footnotesize
\bibliographystyle{aaai}
\bibliography{full,newbib,andrea}
}
\end{document}